\documentclass[11pt]{article}
\usepackage{coling2020}
\usepackage{times}
\usepackage{url}
\usepackage{latexsym}
\usepackage[linesnumbered,algoruled,boxed,lined]{algorithm2e}
\SetKwInOut{Parameter}{parameter}
\usepackage{enumitem}
\usepackage{amsmath}
\DeclareMathAlphabet{\mathpzc}{OT1}{pzc}{m}{it}
\usepackage{graphicx}
\usepackage{multirow}
\usepackage{adjustbox}
\usepackage{hyperref}
\pdfoutput=1

\colingfinalcopy 

\title{Intent Mining from past conversations \\
for conversational agent}

\author{Ajay Chatterjee \\
        Accenture Labs, India \\
  {\tt ajay.chatt03@gmail.com  } \\\And
  Shubhashis Sengupta \\
  Accenture Labs, India \\
  {\tt shubhashis.sengupta@accenture.com} \\}

\date{}

\begin{document}
\maketitle
\begin{abstract}
Conversational systems are of primary interest in the AI community. Organizations are increasingly using chatbot to provide round-the-clock support and to increase customer engagement. Many commercial bot building frameworks follow a standard approach that requires one to build and train an intent model to recognize user input. These frameworks require a collection of user utterances and corresponding intent to train an intent model. Collecting a substantial coverage of training data is a bottleneck in the bot building process. In cases where past conversation data is available, the cost of labeling hundreds of utterances with intent labels is time-consuming and laborious. In this paper, we present an intent discovery framework that can mine a vast amount of conversational logs and to generate labeled data sets for training intent models. We have introduced an extension to the DBSCAN ~\cite{6} algorithm and presented a density-based clustering algorithm ITER-DBSCAN for unbalanced data clustering. Empirical evaluation on one conversation dataset, six intent dataset, and one short text clustering dataset show the effectiveness of our hypothesis. We release the datasets and code for future evaluation at \url{https://github.com/ajaychatterjee/IntentMining} . \footnote{This work is licensed under a Creative Commons Attribution 4.0 International License. License details: http://creativecommons.org/licenses/by/4.0/.}
\end{abstract}

\section{Introduction}
In the past few years, there is a growing community and business interest in conversational systems (chatbots primarily). A key step towards designing a task-oriented conversational model is to identify and understand the intention of a user utterance. An intent in a conversational model maps semantically similar sentences to a high-level abstraction for a chatbot that can generate a similar response or perform an action. For example, ``unable to log-in to the system", ``can not log in", ``facing issue during sign-in" - though linguistically different, are all interpreted as intents related to \textbf{login issue}. The current crop of bot-building frameworks require annotated data for building an Intent model. Many commercial chatbot building frameworks such as Microsoft Azure Bot Service\footnote{\url{https://azure.microsoft.com/en-us/services/bot-service/}}, IBM Watson Assistant\footnote{\url{https://www.ibm.com/cloud/watson-assistant/}} support intent training in a supervised setting. 
The developers and domain experts typically consider past interaction logs between human-human or human-computer as a valuable resource and carry out an extensive manual process of intent labeling. The process of intent discovery and training data creation is by large manual and effort-intensive and carried out by domain experts.

Existing dialog corpora contains pre-defined intent and dialog state defined, and consequently, most of the work~\cite{mrksicetal2015multi,Henderson2014WordBasedDS} ignores intent discovery during conversation design. Previous work~\cite{haponchyk-etal-2018-supervised,p-2016-mixkmeans} on intent identification focuses on clustering single user query/ question using supervised or unsupervised clustering. But the tasks do not consider conversational data. Perkins \shortcite{perkinsyang2019dialog} discusses the realistic complexity of user intent space in a complex domain such as customer support and health care and use the conversational data for clustering and intent induction. But the previous works on intent discovery use pre-decided number of intents as a parameter to group the data. In real-world datasets, estimating the real number of intent is also challenging. To address these problems, we propose a set of data extraction methodology to extract a set of utterances from a conversation. These utterances later will be used for clustering and generation of parallel corpus for intent classification training.


To the best of our knowledge, ours is one of the first efforts to bridge the gap between - 1) research in dialog act tagging~\cite{stolcke-etal-2000-dialogue,kim-etal-2010-classifying,10.1145/3209978.3209997}, 2) state of the art research in Natural Language representation~\cite{28,10.5555/2969442.2969607,DBLP:journals/corr/abs-1803-02893} and, 3) density-based clustering~\cite{6,19} for automatically discovering intents. In this work, we describe an Intent Mining framework that reduces the labeling effort significantly by using two sources of information - the metadata/ short description about conversations and the conversations themselves (Refer to Table 1 for a sample conversation in the helpdesk scenario). In cases where raw conversations are presented without any metadata, we have experimented with different approaches to extract a suitable description for representing the summary/ short description of a conversation. We also experimented with the pre-trained language model (Universal Sentence Encoder \cite{28}) for sentence representation. We use the textual descriptions to cluster conversations into unique groups, using a density-based clustering approach (discussed in section 3.2). Clusters are labeled to generate seed data for each intent. Features extracted from the labeled conversations along with intent labels are used to generate training data and train a statistical classifier. Unlabeled conversations are then labeled by the base classifier on the basis a cut-off confidence score of the model. The final training set can be used to train any supervised classification algorithm. We show that an intent model trained in this manner works with good efficacy and provides decent coverage of intents.
\begin{table}
\centering
\resizebox{11cm}{!} {
\begin{tabular}{|l|p{0.5\linewidth}|}
\hline
 \bf USER &  Hi, is there any way to enable skype recording. \\  
\bf AGENT & Hello USER  \\  
\bf USER & Hi  \\  
\bf AGENT & As I understand, you need recording service to be enabled for Skype for Business. \\  
\hline \bf Issue Description &  User reported unable to record  calls \\ \hline
\end{tabular}
}
{\caption{A sample conversation between a Customer (USER) and a Support Analyst (AGENT) along with Issue description is added by Agent after the conversations in IT Support. The analyst is trying to solve a problem related to Microsoft Skype for Business application.}\label{table1}}
\end{table}

The primary contribution of the work summarized as follows :
\begin{itemize}[noitemsep]
\item We present an intent discovery framework that involves 4 primary steps: 1) extraction of textual utterances from a conversation using a pre-trained domain agnostic Dialog Act Classifier (Data Extraction), 2) automatic clustering of similar user utterances (Clustering),  3) manual annotation of clusters with an intent label (Labeling) and, 4) propagation of the intent labels to the utterances from the previous step, which are not mapped to any cluster (Label Propagation) to generate intent training data from raw conversations.
\item Our work present an effort to generate intent training data for raw conversations. We introduce the dialog intent mining task and present a density based clustering algorithm with novel feature extraction technique.

\item The true class distribution of intents of the real world conversation data is unknown and may contain skewness. Our work presents an effort to automatically discover clusters without any prior knowledge about the intent classes.

\item We study the performance of previous density based clustering algorithm in the intent discovery task. The presented algorithm, ITER-DBSCAN outperforms previous state of the art in terms of intent coverage. 

\end{itemize}

\section{Related Work}
Intent discovery and analysis is a fundamental step to build intelligent task-oriented conversational agents. Intents are a sequence of words which are mapped to predefined categories to comprehend user request. Recent works point to two directions to build quality intent models. Re-using available conversation log, to bootstrap intent model building process \cite{1,2,26,25}. The other is to allow domain experts to build an intent model by working on the model definition, labeling, and evaluation through user interfaces \cite{3}. Our work is at the intersection of these two approaches, in the sense that we mine candidate clusters in an unsupervised way and then allow domain experts to review and label the clusters (Intent Discovery). 

Gathering good quality labeled data for any machine learning process is expensive. There have been significant efforts to reduce labeling effort; including work on clustering\cite{10.1145/2124295.2124342,Xu2017,perkinsyang2019dialog}, semi-supervised learning \cite{16}, active learning \cite{17}, transfer learning \cite{2} and also recently proposed data programming frameworks \cite{18,1}. Semi-supervised, Transfer learning and active learning require seed training data for processing. Clustering is primarily used to collect the initial seed data. Most clustering algorithms fail to discover classes in a highly skewed distribution. We overcome these challenges to obtain labels on noisy data by applying a novel clustering algorithm for seed data collection and subsequently propagating labels to generate high-quality training data. 
Various work focus on using existing chat logs to build intent models. A transfer learning-based system has been proposed \cite{2} to learn from low resource settings. Data programming based \cite{1} systems provide an interface to write labeling function for labeled data generation. However, one underlying assumption of using these methods is that they all require the intents to be known beforehand. This pre-condition is very difficult to meet in real-world cases.

Clustering is also an active research area for pattern mining. Popular algorithms such as centroid based clustering algorithms (K-Means \cite{5}), density-based algorithm (DBSCAN \cite{6}), HDBSCAN \cite{19}), are very useful in practical applications. Although K-Means is very fast and mostly used for clustering, it requires one to define the number of clusters as a parameter to the algorithm. Among the existing clustering approaches, a density-based algorithm particularly DBSCAN (density-based spatial clustering with noise) and its variations, is more efficient for detecting clusters with arbitrary shapes from the noisy dataset where there is no prior knowledge about the number of clusters \cite{24,25}. Many improved versions of this algorithm are also available (such as NG-DBSCAN \cite{4}) to overcome the scalability issues of density-based clustering, but they fail to address the ineffectiveness of density-based approaches in sparse data setting. 

Although density-based clustering has limitations, it is a powerful tool for automatic data exploration and pattern mining.  A key contribution of our work is to provide a better exploration strategy in unbalanced data settings. We search the feature space for different density clusters by adjusting the density definition of DBSCAN algorithm over iterations. This allows us to generate cluster with different densities and hence to find intents with low frequencies from the past chat log. Clusters are explicitly labeled by the expert to collect training data for the intent model. We apply this methodology in the publicly available intent classification dataset with highly skewed class distribution to understand the effectiveness of our clustering algorithm for intent discovery. 

\section{Methodology}
In this section, we describe the methods used for the Intent Mining framework.

\subsection{Feature Engineering}
The following methods are used for extracting features from the Natural language description and conversation data .

\begin{enumerate}
\item \textbf{Pre-trained Sentence Embedding (USE)} : We use pre-trained sentence embedding(Universal Sentence Encoder \cite{28}) without any fine-tuning for the downstream tasks. Here, we pass each short description to the model\footnote{https://tfhub.dev/google/universal-sentence-encoder/4} and extract \textit{1-d} vector representation.
\item \textbf{Dialog Act Classifier} : Dialog Act Classifier \cite{12} is crucial to Natural Language Understanding, as it provides a general representation of speaker's intent, that is not bound to any particular dialog domain. The correct interpretation of the intent behind a speaker's utterance plays an important role in determining the success of the conversation. For example, consider these two utterances - ``Book a flight for me" and ``Can you book a flight". The generic intent of the first utterance is a ``Command" type and where the latter is a ``Question" type, and the domain dependent intent is same for both case, ``book a flight". Understanding different cues of the natural language helps to generate better response. For example, for the first utterance, the dialog system can generate more human-like response,  ``Sure. Please wait for few minutes as I will start the booking process", whereas for the second case it can be more straight forward as ``Alright. Let me start the booking process." 

In the context of our work, we use ATIS Corpus \cite{20}; the dataset contains textual conversations related to Air Travel Information System. Utterances in the conversations are tagged with dialog act types -  ``Information", ``Query", ``Command", ``Greetings", ``Confirmation-Affirmation". Natural language-based features such as part-of-speech of the tokens,  bi-grams of parts-of-speech are extracted from the utterances and a sequence-based classifier (CRF \cite{21}) is trained.  The CRF model is configured with the following parameters - a. training algorithm: lbfgs \cite{22} (Gradient descent using the L-BFGS method), L2 regularization: 0.001. We train a sequence classifier using python-crfsuite\footnote{https://python-crfsuite.readthedocs.io/en/latest/}; to tag utterances in a dialog system with dialog act type.
\end{enumerate}

\subsection{Cluster \& Label}

DBSCAN is a density based clustering non-parametric algorithm, given a set of points, it groups points together that are closely packed (points with many nearby neighbors, high-density area) and marks points  that lie alone in low-density regions (whose nearest neighbors are too far away) as outliers. The primary advantages of density-based clustering one that a) it can automatically find clusters based on the definition of density, b) it can find clusters of arbitrary shape rather than being limited to ``ball-shaped'' ones.  We propose a variation of this algorithm in our work and the primary motivation is driven by the following two research questions - \\

\textbf {Research Question 1 : How to group a set of data points without defining a hard bound on the number of cluster?} 

\textbf {Research Question 2: How to automatically search for clusters with different densities in a sparse data space?} \\

DBSCAN is a popular density-based clustering algorithm that searches for clusters broadly with two parameters - a) Maximum distance and, b) Minimum number of points. In DBSCAN literature, a point is a core point if there exists a threshold number of points within a maximum distance including the core point. All the other points are classified as Noise.

Let, \textbf{X} be a set of points $\{x_\mathrm{1}, x_\mathrm{2}, .., x_\mathrm{i}\}$ to be clustered and the distance between any two points is defined by \textit{D}(,).  \\ \\
Let \textit{S}(\textbf{X}) be a subset of \textbf{X}. And, 
\textit{l} = \textit{D}(p, 0) = \textit{D}(0, p)
such that, \textit{l} is the distance between point p and origin. \\ 
Let, $\mathpzc{N(\mathord{\cdot})}$ be the cardinality of a set. Let, 
${x_\mathrm{i}, x_\mathrm{j}}$ be any two points from the set \textit{S}(\textbf{X}). such that, 
\setlength{\belowdisplayskip}{1pt} \setlength{\belowdisplayshortskip}{1pt}
\setlength{\abovedisplayskip}{1pt} \setlength{\abovedisplayshortskip}{1pt}

\begin{align}
\forall_{\mathrm{i}}\exists_{\mathrm{j}} \textit{D}(x_\mathrm{i}, x_\mathrm{j}) \leq d  \\
\mathpzc{N}(\textit{S}(\textbf{X})) \geq K
\end{align}  

Where,
d is the maximum distance and K is the minimum number of points, according to the definition of DBSCAN.

We formulate that, there also exists a subset \textit{P}(\textbf{X}) and let ${x_\mathrm{i}, x_\mathrm{j}}$ be any two points in it. Then, 

\begin{align}
\exists_{\mathrm{\textit{P}(\textbf{X})}}\forall_{\mathrm{i}}\exists_{\mathrm{j}} \textit{D}(x_\mathrm{i}, x_\mathrm{j}) \leq d + \delta d \\
\mathpzc{N}(\textit{P}(\textbf{X})) \geq K^\mathrm{'} 
\end{align}

where, $K > K^\mathrm{'}$. Equations 3 and 4 essentially tell that less frequent classes in the dataset can be found by increasing the distance value constraint and propose a minimum number points constraint for cluster discovery to tackle unbalanced data distribution. \\

We modify the DBSCAN algorithm, naming it ITER-DBSCAN, to work with datasets having imbalanced class distribution (Refer to Algorithm 1). The algorithm runs iteratively to search for clusters with high-density regions to low-density regions. The low-density region search is controlled by two parameters ``max-distance''and ``min-points''. ``max-distance'' parameters controls what is maximum distance to consider two items belongs to same group  and ``min-points'' controls what is minimum number of items in a group to qualify it as a cluster. We use cosine-distance for calculating distance between points. 

\begin{algorithm}[H]
  \KwIn{A set of Textual utterances(data-points)}
  \Parameter{featuretransformer, initial-min-distance, initial-number-of-points, delta-min-distance, delta-number-of-points, min-points, max-iteration}
  \KwOut{Data-points with cluster label }

  current-minimum-distance=initial-min-distance\;
  current-number-of-points=initial-number-of-points\;
  iteration=1\;
  \While{iter $\leq$ max-iteration}{
    \If{current-number-of-points == min-points}{
      break\;
      }
    \tcc{compute feature representation of the data points with the featuretransformer method}
    feature-vector=featuretransformer(data-points) \;
    \BlankLine
    Run DBSCAN Algorithm with current-minimum-distance, current-number-of-points and feature-vector\;
    current-data-points = get data points marked as noisy points\;
    set data-points with current-data-points\;
    current-minimum-distance = current-minimum-distance - delta-min-distance\;
    current-number-of-points = current-number-of-points - delta-number-of-points \;
    iteration = iteration + 1 \;
  }
  \caption{ITER-DBSCAN}
\end{algorithm}

\textbf{Parameters}: ITER-DBSCAN parameters are descried below,

\begin{itemize}[noitemsep]
\item \textbf{data-points}: The primary input to the algorithm is a set of data-points (textual data) for clustering.
\item \textbf{featuretransformer}: Transformer function to convert the textual data into feature representation\footnote{In this work the feature representation is referred as numerical feature.}.
\item \textbf{initial-min-distance}: Initial distance value for creating cluster.
\item \textbf{initial-number-of-points}: Initial number of points in a group for cluster validation.
\item \textbf{delta-min-distance}: Single distance value is not enough to cluster sparse dataset, at each iteration the distance value is increased by delta-min-distance parameter to search for new cluster.
\item \textbf{delta-number-of-points}: Minimum number of points parameter is decreased by delta-number-of-points parameter at each iteration for finding low density cluster.
\item \textbf{min-points}: Iteration is terminated when the minimum number of points for cluster creation reaches min-points.
\item \textbf{max-iteration}: max-iteration is the maximum number of times algorithm runs and updates delta-min-distance and delta-number-of-points for cluster discovery.
\end{itemize}

\subsection{Label Propagation}
Our clustering approach provides a set of labeled conversation-intent pair and a set of unlabeled conversations, as the density-based clustering might not group all the data points. A statistical classifier such as Logistic Regression\footnote{Other statistical classifier or neural network-based model might provide better accuracy, but this part is out of the scope of our current research} learns a mapping function between labelled conversations and intents. The trained classifier propagates the intent to the unlabelled conversations to generate a final intent classification training dataset.

\subsection{Approaches for Description Extraction form Conversation}
In industrial service desk scenario, the metadata or description about the conversation is added later by the service agent after the issue is resolved and may not available in many cases. In this section, we describe two methods to extract textual descriptions from the raw conversation logs which can then be fed into our clustering model - 

\begin{itemize}[noitemsep]
\item The agent answering to the service call always clarifies the intent with the user. Therefore, we can extract all the question asked by the Agent during the conversation with Dialog Act Classifier model (an SVM classifier created using trained Fasttext word embeddings as feature) and apply our clustering and label propagation approach to find different set of questions asked by the agent. We mark a special type of questions asked by the agent is ``intent\_clarfication'' to clarify the intent. For example - ``As I understand you need recording service to be enabled for Skype for Business'' (Refer to Table 1)  where Agent clarifies the request with the request with user. We can extract this sentence for Short description of the conversation.
\item We can also extract top-3 user utterances of ``Information'' or ``Question" type using Dialog Act Classifier model. This utterance set can be also used for representing a short description about the conversation. This design choice is made from the observation that a user informs about her queries in the top few messages and DAC model filters some of the unrelated utterances (such as greetings and command type) leaving behind the ones that can be used in our purpose. COMMAND type utterance removal is a special case, since in our conversation dataset users request rather than command agents for help. But in other scenario, we might need to add COMMAND type utterances for representing short description of the conversation. 

\end{itemize}
\section{Data}
Existing task oriented dialog corpora such as  MultiWOZ dataset~\cite{budzianowski2018multiwoz,ramadan2018large,eric2019multiwoz}, Microsoft Dialog Dataset~\cite{li2018microsoft,li2016user}, ATIS~\cite{hakkani-tur2016multi} comprise of dialog intent defined in narrow domain like Train, Restaurant, car booking. Recently, Perkins \shortcite{perkinsyang2019dialog} published a curated complex human-human conversation dataset gathered from Twitter airline customer support. The tweets comprise conversations between customer support agents of some airline companies and their customers. The conversations constitutes various topics for intent mining task. We also consider various open-source short text dataset and evaluate the generalization of our algorithm. In 
, we present the overview of the datasets. Table 2 also presents the intent distribution i.e., the maximum and the minimum number of utterances pertaining to one intent.

\begin{table}[h]
\centering
\resizebox{11cm}{!} {
\begin{tabular}{|l|l|l|l|l|}
\hline
\textbf{Dataset} & \textbf{Utterances} & \textbf{Intents} & \textbf{Max Intent count} & \textbf{Min Intent count} \\ \hline
Airlines Twitter Conversation & 491 & 14 & 107 & 10 \\ \hline
FinanceData & 10003 & 77 & 187 & 35 \\ \hline
AskUbuntuCorpus & 162 & 5 & 57 & 8 \\ \hline
ChatbotCorpus & 206 & 2 & 128 & 78 \\ \hline
WebApplicationCorpus & 88 & 7 & 23 & 5 \\ \hline
ATIS & 4972 & 17 & 3666 & 6 \\ \hline
Personal Assistant & 25312 & 64 & 1218 & 171 \\ \hline
Stackoverflow Dataset & 20000 & 20 & 1000 & 1000 \\ \hline
\end{tabular}
}
{\caption{Overview of the datasets used in the evaluation.}\label{table2}}
\end{table}

\subsection{Conversation Dataset}
The Airlines Twitter Conversation dataset~\cite{perkinsyang2019dialog} is a human-human conversation dataset\footnote{https://github.com/asappresearch/dialog-intent-induction} related to user queries posted on various topics on Twitter about airline-related travel. We extract a short description from these conversations, discussed in section 3.4.

\subsection{Dialog Intent and Short Text Dataset}
The Finance dataset~\cite{Casanueva2020} contains various online banking queries annotated with their corresponding intents\footnote{https://github.com/PolyAI-LDN/task-specific-datasets} publised by PolyAI team. The AskUbuntuCorpus, ChatbotCorpus and WebApplicationCorpus, this three corpora~\cite{braun-EtAl:2017:SIGDIAL} collected from StackExchange and Telegram Chatbot contains utterances with intent labels\footnote{https://github.com/sebischair/NLU-Evaluation-Corpora}. The ATIS dataset\footnote{https://www.kaggle.com/hassanamin/atis-airlinetravelinformationsystem}, which provides large number of messages and their associated intents, is useful for intent discovery/ classification problems including chatbots. Personal Assistant is another large scale dataset~\cite{XLiu.etal:IWSDS2019} consisting of messages posted by a personal assistant. The dataset\footnote{https://github.com/xliuhw/NLU-Evaluation-Data} contains 25K+ messages with 64 intent label. Stackoverflow Dataset\footnote{https://github.com/jacoxu/StackOverflow} is a short text dataset used for classification and clustering of extracted queries from StackOverflow website.

\section{Experiments}
In this section, we evaluate ITER-DBSCAN on the 8 datasets discussed in Section 4. We compare ITER-DBSCAN algorithm with state-of-the-art density-based clustering algorithms such as DBSCAN and HDBSCAN. For the conversation dataset, we also evaluate our results with a recently published multi-view clustering, AV-KMeans Algorithm. We use USE sentence embedding method to convert the natural language to numerical feature.

\subsection{Metrics}
We use standard clustering metrics to evaluate the algorithms. We adapt metrics such as precision, recall, F1 score, and unsupervised clustering accuracy from the work of Perkins~\shortcite{perkinsyang2019dialog} to compare our results for the conversation datasets. To evaluate the short text datasets, we primarily use two metrics: a) Normalized Mutual Information~\cite{10.1145/1553374.1553511}, b) Adjusted Rand Index~\cite{Hubert1985}. 

\textbf{Normalized Mutual Information(NMI):} NMI is designed as a combined measure for the accuracy of clustering and the total number of clusters. NMI is an entropy based metric that computes the amount of common information between two partitions - \\
\begin{equation}
    \mathrm{NMI}={\dfrac{2 * I(Y;C)}{H(Y)+H(C)}}
\end{equation}
    
where, Y is class labels, C is cluster labels, H(.) is Entropy and I(Y;C) is Mutual information between Y and C.

\textbf{Adjusted Rand Index(ARI):} ARI computes a similarity measure between two clusterings by considering all pairs of samples and counting pairs that are assigned in the same or different clusters in the predicted and true clusterings. ARI is an improved version of Rand
Index, which is defined as follows:

\begin{equation}
\mathrm{ARI}=\frac{\sum_{i j}\left(\begin{array}{l}{n_{i j}} \\ {2}\end{array}\right)-\left[\sum_{i}\left(\begin{array}{l}{a_{i}} \\ {2}\end{array}\right) \sum_{j}\left(\begin{array}{l}{b_{j}} \\ {2}\end{array}\right)\right] /\left(\begin{array}{l}{n} \\ {2}\end{array}\right)}{\frac{1}{2}\left[\sum_{i}\left(\begin{array}{l}{a_{i}} \\ {2}\end{array}\right)+\sum_{j}\left(\begin{array}{l}{b_{j}} \\ {2}\end{array}\right)-\left[\sum_{i}\left(\begin{array}{l}{a_{i}} \\ {2}\end{array}\right) \sum_{j}\left(\begin{array}{l}{a_{j}} \\ {2}\end{array}\right)\right] /\left(\begin{array}{l}{n} \\ {2}\end{array}\right)\right.}
\end{equation}

The range of AMI and NMI is from 0 to 1, a larger value indicates a higher agreement between ground truth and the predicted partition for the dataset.

\subsection{Parameter Settings}
We evaluate DBSCAN, HDBSCAN and ITER-DBSCAN on a wide variety of parameters. We use the scikit-learn implementation of DBSCAN\footnote{https://scikit-learn.org/stable/modules/generated/sklearn.cluster.DBSCAN.html} and HDBSCAN\footnote{https://github.com/scikit-learn-contrib/hdbscan}. We generate a wide range based on the combination of the parameters described below. All the other parameters are kept at default according to the implementation. We use cosine distance function for evaluating the clustering algorithms. To evaluate the results better, we keep the minimum cluster size as 3 for all the density based clustering algorithm.\\
	
\begin{itemize}
\item \textbf{DBSCAN:} ~For evaluaton of DBSCAN, we select the minimum distance parameter between a range 0.09 to 0.40 with a change of 0.01 (Example: [0.09, 0.10, 0.11, .., 0.40]). We also configure the minimum sample parameter between a range 3 to 20 with a change of 1.\\
\item \textbf{HDSCAN:} We configure the minimum cluster size parameter between a range of 3 to 15 with a change of 1. We set the minimum samples parameter between a range 3 to 15 with a change of 1.\\
\item \textbf{ITER-DBSCAN}: We set a range of values for three parameters of ITER-DBSCAN for evaluation. We use five initial distance value 0.09, 0.12, 0.15, 0.20, 0.30. We set the maximum iteration and initial minimum sample parameters between a range 10 to 15 and 10 to 25 respectively, with a change step of 1. We keep the other parameters constant - such as, delta-min-distance as 0.01, delta-number-of-points as 1, minimum points as 3. 
\end{itemize}

\subsection{Results}
In Table 3, we present the evaluation of our algorithm on Twitter airline conversation dataset(TwACS) with DBSCAN and AV-KMeans. We use 4 evaluation metrics adapted from the work of Perkins~\shortcite{perkinsyang2019dialog}. The HDBSCAN algorithm did not find any clusters for this dataset, hence not reported. We evaluate the dataset with the top-3 utterances extracted from conversation, with (and without) dialog act classifier based feature extraction. We report the effectiveness of feature extraction methodology in Table 3. \\

In Table 4, we present our algorithm ITER-DBSCAN results as a comparison to DBSCAN and HDBSCAN.  For each dataset, we describe the total number of intents and the number of intents the algorithm identified. We also present the Normalized Mutual information score and Adjusted Rand Score for clustering evaluation. In most of the task, our algorithm achieves state-of-the-art results on the intent discovery and different clustering metrics.

\begin{table}[htbp]
\centering
\resizebox{11cm}{!} {
\begin{tabular}{|c|l|l|l|l|l|}
\hline
Corpus & \multicolumn{1}{c|}{Algorithm} & \multicolumn{1}{c|}{Precision} & \multicolumn{1}{c|}{Recall} & \multicolumn{1}{c|}{F1} & \multicolumn{1}{c|}{ACC} \\ \hline
\multirow{4}{*}{TwACS} & DBSCAN & 31.8 & \textbf{65.4} & 42.8 & 31.8 \\ \cline{2-6} 
 & AV-KMEANS & \textbf{48.9} & 43.8 & 46.2 & 39.9 \\ \cline{2-6} 
 & ITER-DBSCAN (without feature extraction) & 42.7 & 48.3 & 47.4 & 37.5\\ \cline{2-6} 
 & ITER-DBSCAN (with feature extraction)& 48.5 & 54.4 & \textbf{51.3} & \textbf{48.5} \\ \hline
\end{tabular}
}
{\caption{Experiment result of Twitter airline conversation dataset}\label{table3}}
\end{table}

\begin{table}[htbp]
\centering
\resizebox{11cm}{!} {
\begin{tabular}{|c|l|l|l|l|l|}
\hline
\textbf{CorpusName} & \multicolumn{1}{c|}{\textbf{Algorithm}} & \multicolumn{1}{c|}{\textbf{Total Intents}} & \multicolumn{1}{c|}{\textbf{Intents Found}} & \multicolumn{1}{c|}{\textbf{NMI}} & \multicolumn{1}{c|}{\textbf{ARI}} \\ \hline
\multirow{3}{*}{\textbf{AskUbuntuCorpus}} & DBSCAN & 5 & 3 & 0.35 & 0.23 \\ \cline{2-6} 
 & HDBSCAN & 5 & 4 & 0.44 & 0.33 \\ \cline{2-6} 
 & ITER-DBSCAN & 5 & 4 & \textbf{0.51} & \textbf{0.45} \\ \hline
\multirow{3}{*}{\textbf{ATIS}} & DBSCAN & 17 & 13 & 0.28 & 0.26 \\ \cline{2-6} 
 & HDBSCAN & 17 & 11 & 0.24 & 0.24 \\ \cline{2-6} 
 & ITER-DBSCAN & 17 & 14 & \textbf{0.55} & \textbf{0.66} \\ \hline
\multirow{3}{*}{\textbf{ChatbotCorpus}} & DBSCAN & 2 & 2 & 0.59 & 0.61 \\ \cline{2-6} 
 & HDBSCAN & 2 & 2 & 0.59 & 0.61 \\ \cline{2-6} 
 & ITER-DBSCAN & 2 & 2 & \textbf{0.63} & \textbf{0.68} \\ \hline
\multirow{3}{*}{\textbf{FinanceData}} & DBSCAN & 77 & 76 & 0.44 & 0.17 \\ \cline{2-6} 
 & HDBSCAN & 77 & 75 & 0.53 & 0.31 \\ \cline{2-6} 
 & ITER-DBSCAN & 77 & 77 & \textbf{0.79} & \textbf{0.6} \\ \hline
\multirow{3}{*}{\textbf{PersonalAssistant}} & DBSCAN & 64 & 64 & 0.45 & 0.25 \\ \cline{2-6} 
 & HDBSCAN & 64 & 64 & 0.64 & 0.48 \\ \cline{2-6} 
 & ITER-DBSCAN & 64 & 64 & \textbf{0.65} & \textbf{0.5} \\ \hline
\multirow{3}{*}{\textbf{Stackoverflow}} & DBSCAN & 20 & 20 & 0.48 & 0.34 \\ \cline{2-6} 
 & HDBSCAN & 20 & 20 & \textbf{0.72} & 0.62 \\ \cline{2-6} 
 & ITER-DBSCAN & 20 & 20 & 0.71 & \textbf{0.63} \\ \hline
\multirow{3}{*}{\textbf{WebApplicationsCorpus}} & DBSCAN & 7 & 4 & 0.33 & 0.22 \\ \cline{2-6} 
 & HDBSCAN & 7 & 4 & 0.32 & 0.23 \\ \cline{2-6} 
 & ITER-DBSCAN & 7 & 5 & \textbf{0.45} & \textbf{0.39} \\ \hline
\end{tabular}
}
{\caption{Experiment result of Intent and short text clustering datasets.}\label{table4}}
\end{table}

\subsection{Parameter Study}
In this section, we study the growth of number of clusters to identify different number of intents. We present the result of all datasets in figure 1. We plot the change of intent counts in x axis and the change of number of clusters in y axis. We also study the effect of parameter configurations for the Finance Dataset in figure 2, and how it changes the number of intents and clusters. In x-axis of the figure 2, we plot the difference between initial minimum distance and maximum iteration which can be regarded as the minimum possible cluster size. In y-axis of the figure 2, we plot the number of intents (left) and number of clusters (right). In figure 2, we lay the change of intent count with respect to different initial distance. The plot also shows that the number of clusters decreases as the difference between as minimum cluster size increases and initial distance between 0.12 to 0.20 provides better stability in discovering intents. So, in practice we can use this two parameters to balance between the number of clusters and the coverage of the intents.
\begin{figure}[!ht]
\centering
  \includegraphics[width=0.8\linewidth]{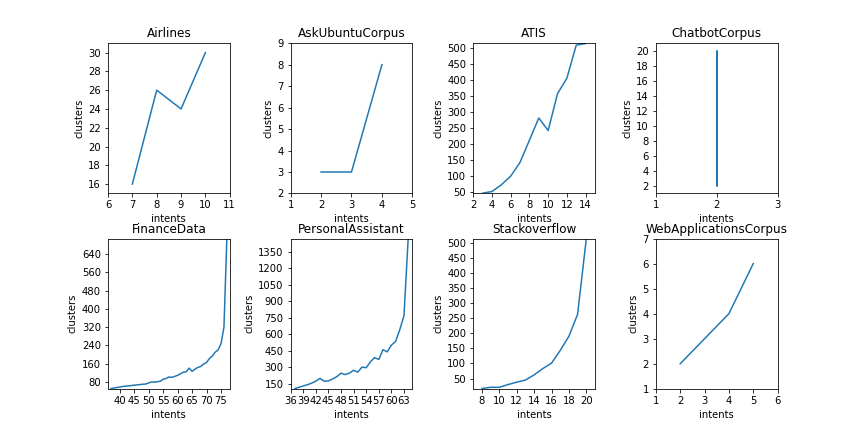}
  \caption{Growth of number of clusters with respect to intent.}
  \label{fig:result}
\end{figure}

\begin{figure}[!ht]
\centering
  \includegraphics[width=0.8\linewidth]{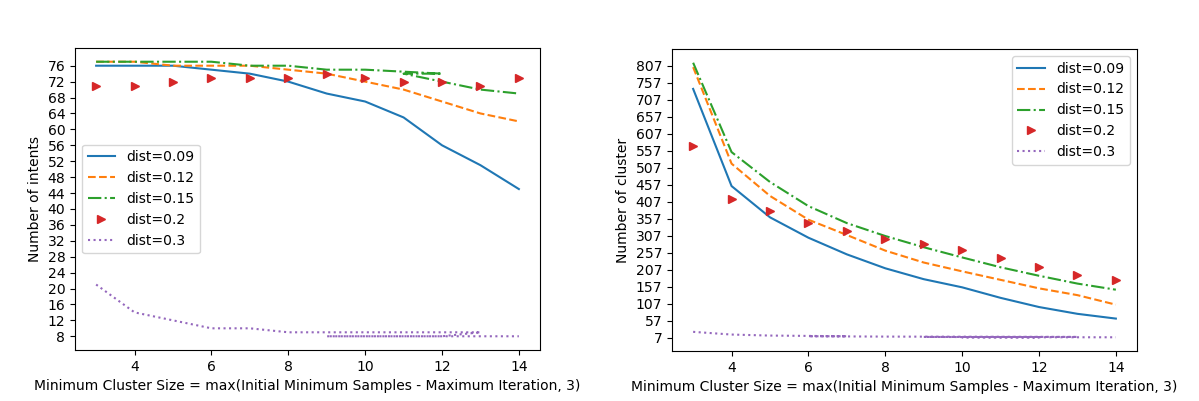}
  \caption{Change of Intent and cluster count with respect to different parameter configuration.}
  \label{fig:paramresult}
\end{figure}
\subsection{Implantation Details and Time Cost Analysis}
Density based clustering algorithm uses local topological structure to create meaningful clusters.  Time cost increases with data dimension and number of points~\cite{10.1145/2723372.2737792}. To overcome this complexity, we partition large dataset into distinct subsets and apply our algorithm to this subsets in parallel. We use K-Means clustering for generating subsets of data when the dataset size is more than 10K. We use the following function~(5) - to set the number of clusters for K-Means algorithm :
\begin{equation}   
Number Of Cluster = Max(data\_size/10000, 3)
\end{equation}

In figure 3, we plot the time taken by density-based clustering algorithms to process the datasets of section 4. Due to parallelization, the time complexity of our algorithm becomes almost linear with dataset size (after reaching volume of 10K). 

\begin{figure}[!ht]
\centering
  \includegraphics[width=0.5\linewidth]{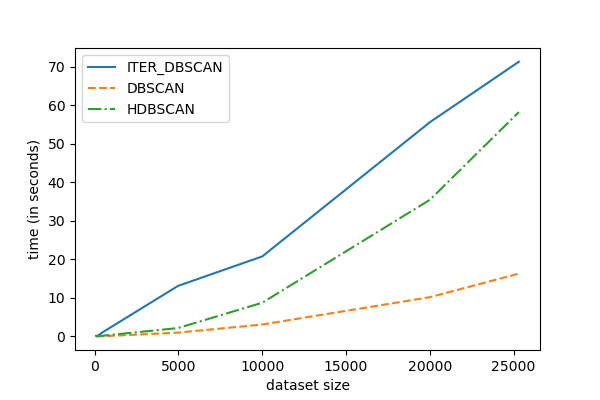}
  \caption{Time complexity analysis of algorithms.}
  \label{fig:paramresult}
\end{figure}

\section{Conclusion and Feature Work}
In this work, we presented a framework that can cluster textual data using a state-of-art sentence representation method with our algorithm. That provides better intent discovery for complex conversation datasets and short text datasets. Our algorithms are able to identify intents from imbalanced dataset with greater accuracy than previous state of the art algorithms. We also presented a feature extraction method using Dialog Act Classification model to extract a short description from conversations for intent mining task.

In future, we would like to extend our work by incorporating various other features from conversations, such as different form of questions asked by the agent to resolve a functional task and other linguistic features, for improved clustering. We would also like to explore the use of the neural network to learn generic conversation representation from chat logs for better feature representation.

\clearpage
\bibliographystyle{coling}
\bibliography{coling2020}

\end{document}